%% file: main.tex
\documentclass[10pt,twocolumn,letterpaper]{article}

\usepackage[pagenumbers]{cvpr} 

\input{preamble}

\definecolor{cvprblue}{rgb}{0.21,0.49,0.74}
\usepackage[pagebackref,breaklinks,colorlinks,citecolor=cvprblue]{hyperref}

\usepackage{graphicx}
\usepackage{amsmath}
\usepackage{amssymb}
\usepackage{booktabs}
\usepackage{caption}

\usepackage[dvipsnames]{xcolor}

\usepackage{makecell}

\usepackage{marvosym}

\def\paperID{***} 
\def\confName{CVPR}
\def\confYear{2024}

\title{\textit{FreeU:} Free Lunch in Diffusion U-Net}

\author{Chenyang Si
\quad
Ziqi Huang
\quad
Yuming Jiang
\quad
Ziwei Liu\textsuperscript{\Letter}\\
S-Lab, Nanyang Technological University
\quad
\\
{\tt\small \{chenyang.si, ziqi002, yuming002, ziwei.liu\}@ntu.edu.sg}\\
}

\begin{document}


\twocolumn[{%
    \renewcommand\twocolumn[1][]{#1}%
    \vspace{-3em}
    \maketitle
    \begin{center}
        \centering
        \includegraphics[width=0.99\textwidth]{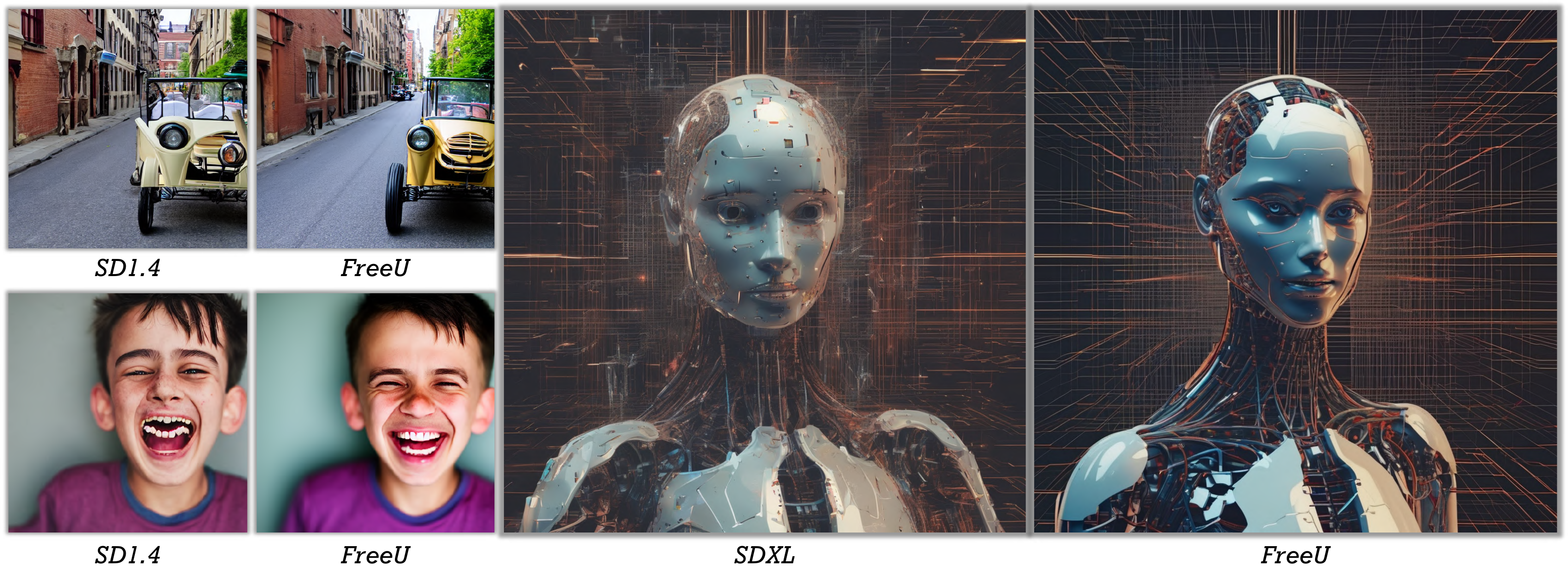}
        \captionof{figure}{
        We propose \textbf{\textit{FreeU}}, a method that substantially improves diffusion model sample quality at no costs: no training, no additional parameter introduced, and no increase in memory or sampling time. 
        }
        \label{fig:teaser}
        \vspace{5pt}
    \end{center}%
    }]

\input{sec/0_abstract}    
\input{sec/1_intro}

\input{sec/2_method}

\input{sec/3_experiment}

\input{sec/4_conclusion}

{
    \small
    \bibliographystyle{ieeenat_fullname}
    \bibliography{egbib}
}


\end{document}

%% file: preamble.tex
\usepackage[dvipsnames]{xcolor}


%% file: sec/0_abstract.tex
\begin{abstract}
In this paper, we uncover the untapped potential of diffusion U-Net, which serves as a ``free lunch'' that substantially improves the generation quality on the fly. We initially investigate the key contributions of the U-Net architecture to the denoising process and identify that its main backbone primarily contributes to denoising, whereas its skip connections mainly introduce high-frequency features into the decoder module, causing the network to overlook the backbone semantics. 
Capitalizing on this discovery, we propose a simple yet effective method—termed ``\textbf{FreeU}'' — that enhances generation quality without additional training or finetuning. Our key insight is to strategically re-weight the contributions sourced from the U-Net’s skip connections and backbone feature maps, to leverage the strengths of both components of the U-Net architecture. Promising results on image and video generation tasks demonstrate that our FreeU can be readily integrated to existing diffusion models, \eg, Stable Diffusion, DreamBooth, ModelScope, Rerender and ReVersion, to improve the generation quality with only a few lines of code. \textbf{All you need is to adjust two scaling factors during inference.}
Project page: \url{https://chenyangsi.top/FreeU/}.
\end{abstract}

%% file: sec/1_intro.tex
\section{Introduction}
\label{sec:intro}



\begin{figure*}
    \centering
    \begin{minipage}{0.57\textwidth}
        \centering
        \includegraphics[width=0.94\textwidth]{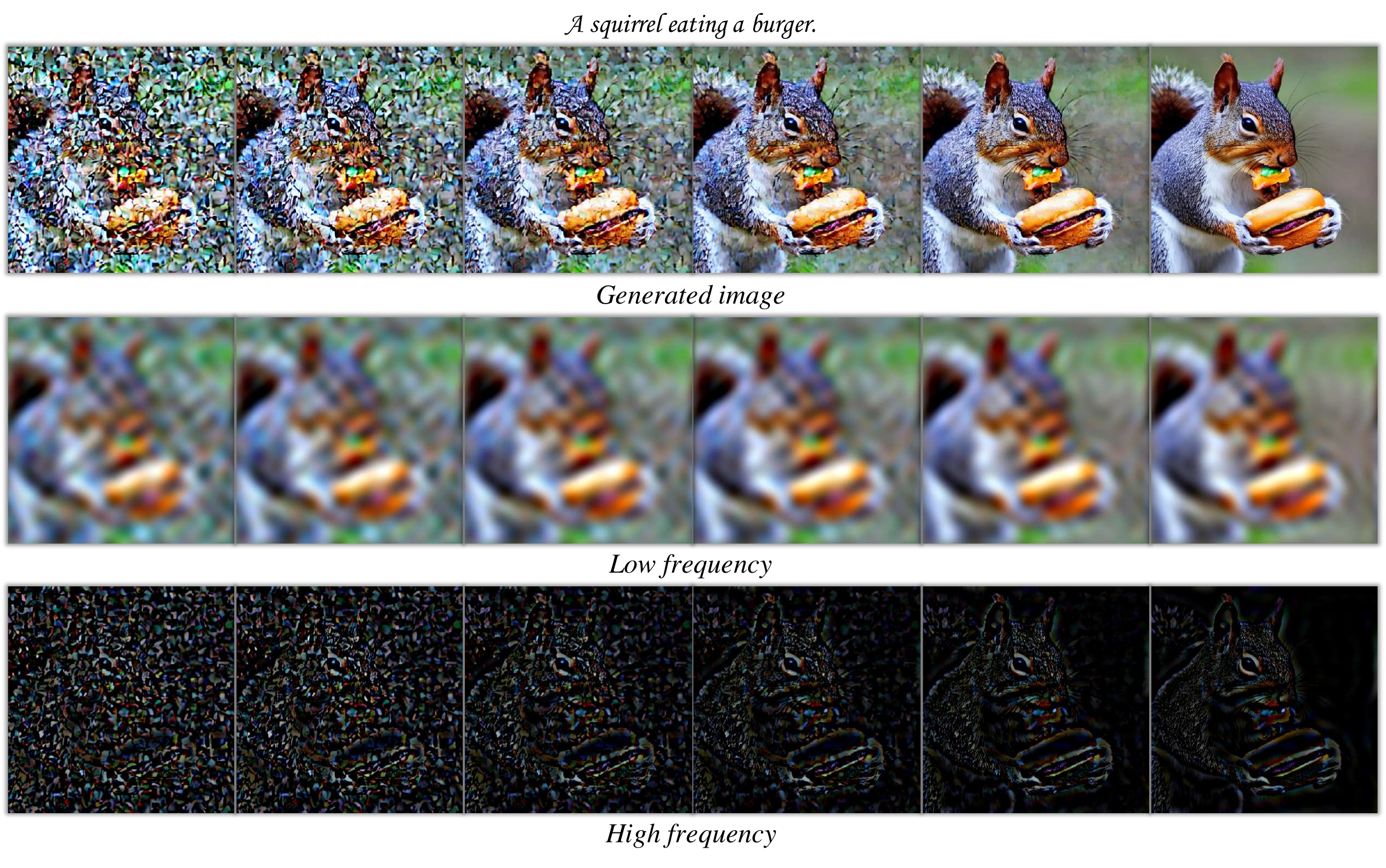}
        \caption{\textbf{The denoising process.} The top row illustrates the image's progressive denoising process across iterations, while the subsequent two rows display low-frequency and high-frequency components after the inverse Fourier Transform, matching each step. It's evident that low-frequency components change slowly, whereas high-frequency components exhibit more significant variations during the denoising process.}
        \label{fig:fig_high_low_img_step}
    \end{minipage}\hfill
    \begin{minipage}{0.42\textwidth}
        \centering
        \includegraphics[width=1.0\linewidth]{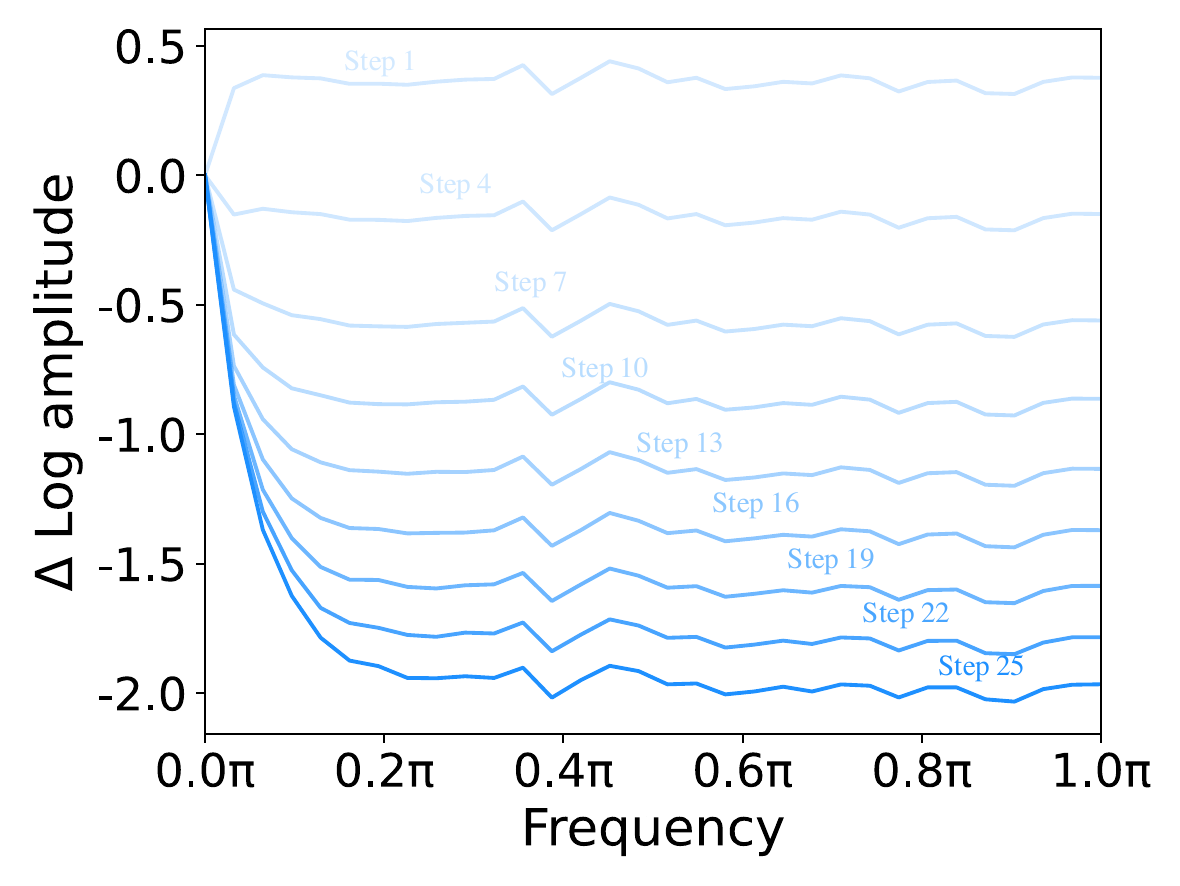}
        \caption{\textbf{Relative log amplitudes of Fourier with variations of the backbone scaling factor $b$.} Increasing in $b$ correspondingly results in a suppression of high-frequency components in the images generated by the diffusion model.}
        \label{fig:fig_each_step_baseline}
    \end{minipage}
\end{figure*}

Diffusion probabilistic models, a cutting-edge category of generative models, have become a focal point in the research landscape, particularly for tasks related to computer vision \cite{ho2020ddpm, dhariwal2021beatgan, rombach2022ldm, ramesh2022dalle2, esser2021imagebart, gu2022vqdiffusion, nichol2021glide, saharia2022imagen, gal2022textualinversion, kumari2022customdiffusion, kawar2022imagic}. 
Distinct from other classes of generative models~\cite{kingma2013vae, wang2021rethinking, goodfellow2014gan, mirza2014cgan, brock2018biggan, karras2018pggan, karras2019stylegan1, karras2020stylegan2, karras2021stylegan3, van2017vqvae, esser2021vqgan} such as Variational Autoencoder (VAE)~\cite{kingma2013vae}, Generative Adversarial Networks (GANs)~\cite{goodfellow2014gan, mirza2014cgan, brock2018biggan, karras2018pggan, karras2019stylegan1, karras2020stylegan2, karras2021stylegan3}, and vector-quantized approaches~\cite{van2017vqvae, esser2021vqgan},
diffusion models introduce a novel generative paradigm. These models employ a fixed Markov chain to map the latent space,  facilitating intricate mappings that capture latent structural complexities within a dataset. Recently, its impressive generative capabilities, ranging from the high level of details to the diversity of the generated examples, have fueled groundbreaking advancements in a variety of computer vision applications such as image synthesis~\cite{ho2020ddpm, rombach2022ldm, saharia2022imagen}, image editing~\cite{avrahami2022blended, choi2021ilvr, meng2021sdedit, huang2023collaborative}, image-to-image translation~\cite{choi2021ilvr, saharia2022palette, wang2022piti}, and text-to-video generation~\cite{hong2022cogvideo, luo2023videofusion, singer2022makeavideo, wu2022tuneavideo, blattmann2023videoldm, he2022lvdm, zhang2023show1, wang2023lavie}.

The diffusion models are comprised of the \textit{diffusion process} and the \textit{denoising process}. During the \textit{diffusion process}, Gaussian noise is gradually added to the input data and eventually corrupts it into approximately pure Gaussian noise. During the \textit{denoising process}, the original input data is recovered from its noise state through a learned sequence of inverse diffusion operations. Usually, a U-Net is trained to iteratively predict the noise to be removed at each denoising step. Existing works focus on utilizing pre-trained diffusion U-Nets for downstream applications, while the internal properties of the diffusion U-Net, remain largely under-explored.


Beyond the application of diffusion models, in this paper, we are interested in investigating the effectiveness of diffusion U-Net for the denoising process. To better understand the denoising process, we first present a paradigm shift toward the Fourier domain to perspective the generated process of diffusion models, a research area that has received limited prior investigation. As illustrated in Fig.~\ref{fig:fig_high_low_img_step}, the uppermost row provides the progressive denoising process, showcasing the generated images across successive iterations. The subsequent two rows exhibit the associated low-frequency and high-frequency spatial domain information after the inverse Fourier Transform, aligning with each respective step.

Evident from Fig.~\ref{fig:fig_high_low_img_step} is the gradual modulation of low-frequency components, exhibiting a subdued rate of change, while their high-frequency components display more pronounced dynamics throughout the denoising process. These findings are further corroborated in Fig.\ref{fig:fig_each_step_baseline}. This can be intuitively explained: 1) Low-frequency components inherently embody the global structure and characteristics of an image, encompassing global layouts and smooth color. 
These components encapsulate the foundational global elements that constitute the image's essence and representation. Its rapid alterations are generally unreasonable in denoising processes. Drastic changes to these components could fundamentally reshape the image's essence, an outcome typically incompatible with the objectives of denoising processes. 2) Conversely, high-frequency components contain the rapid changes in the images, such as edges and textures. These finer details are markedly sensitive to noise, often manifesting as random high-frequency information when noise is introduced to an image. Consequently, denoising processes need to expunge noise while upholding indispensable intricate details.

In light of these observations between low-frequency and high-frequency components during the denoising process, we extend our investigation to ascertain the specific contributions of the U-Net architecture within the diffusion framework. In each stage of the U-Net decoder, the skip features from the skip connection and the backbone features are concatenated together. Our investigation reveals that the main backbone of the U-Net primarily contributes to denoising. Conversely, the skip connections are observed to introduce high-frequency features into the decoder module. These connections propagate fine-grained semantic information to make it easier to recover the input data. However, an unintended consequence of this propagation is the potential weakening of the backbone's inherent denoising capabilities during the inference phase. This can lead to the generation of abnormal image details, as illustrated in the first row of Fig.~\ref{fig:teaser}.

Building upon this revelation, we propel forward with the introduction of a novel strategy, denoted as ``\textbf{FreeU}'', which holds the potential to improve sample quality without necessitating the computational overhead of additional training or fine-tuning. During the inference stage, we instantiate two specialized modulation factors designed to balance the feature contributions from the U-Net architecture's primary backbone and skip connections. The first, termed the backbone feature factors, aims to amplify the feature maps of the main backbone, thereby bolstering the denoising process. However, we find that while the inclusion of backbone feature scaling
factors yields significant improvements, it can occasionally lead to an undesirable oversmoothing of textures. To mitigate this issue, we introduce the second factor, skip feature scaling factors, aiming to alleviate the problem of texture oversmoothing.

Our FreeU framework exhibits seamless adaptability when integrated with existing diffusion models, encompassing applications like text-to-image generation and text-to-video generation. We conduct a comprehensive experimental evaluation of our approach, employing Stable Diffusion~\cite{rombach2022ldm}, DreamBooth~\cite{ruiz2022dreambooth}, ReVersion~\cite{huang2023reversion}, ModelScope~\cite{luo2023videofusion}, and Rerender~\cite{yang2023rerender} as our foundational models for benchmark comparisons. By employing FreeU during the inference phase, these models indicate a discernible enhancement in the quality of generated outputs. The visualization illustrated in Fig.~\ref{fig:teaser} substantiates the efficacy of FreeU in significantly enhancing both intricate details and overall visual fidelity within the generated images. Our contributions are summarized as follows:

\begin{itemize}
    \setlength\itemsep{0em}
    \item We investigate and uncover the potential of U-Net architectures for denoising within diffusion models and identify that its main backbone primarily contributes to denoising, whereas its skip connections introduce high-frequency features into the decoder module.
    \item We further introduce a simple yet effective method, denoted as ``\textbf{FreeU}'', which enhances U-Net's denoising capability by leveraging the strengths of both components of the U-Net architecture. It substantially improves the generation quality without requiring additional training or fine-tuning.

    \item The proposed FreeU framework is versatile and seamlessly integrates with existing diffusion models. We demonstrate significant sample quality improvement across various diffusion-based methods, showing the effectiveness of FreeU at no extra cost.
    
\end{itemize}

%% file: sec/2_method.tex
\begin{figure*}[t]
  \centering
   \includegraphics[width=0.85\textwidth]{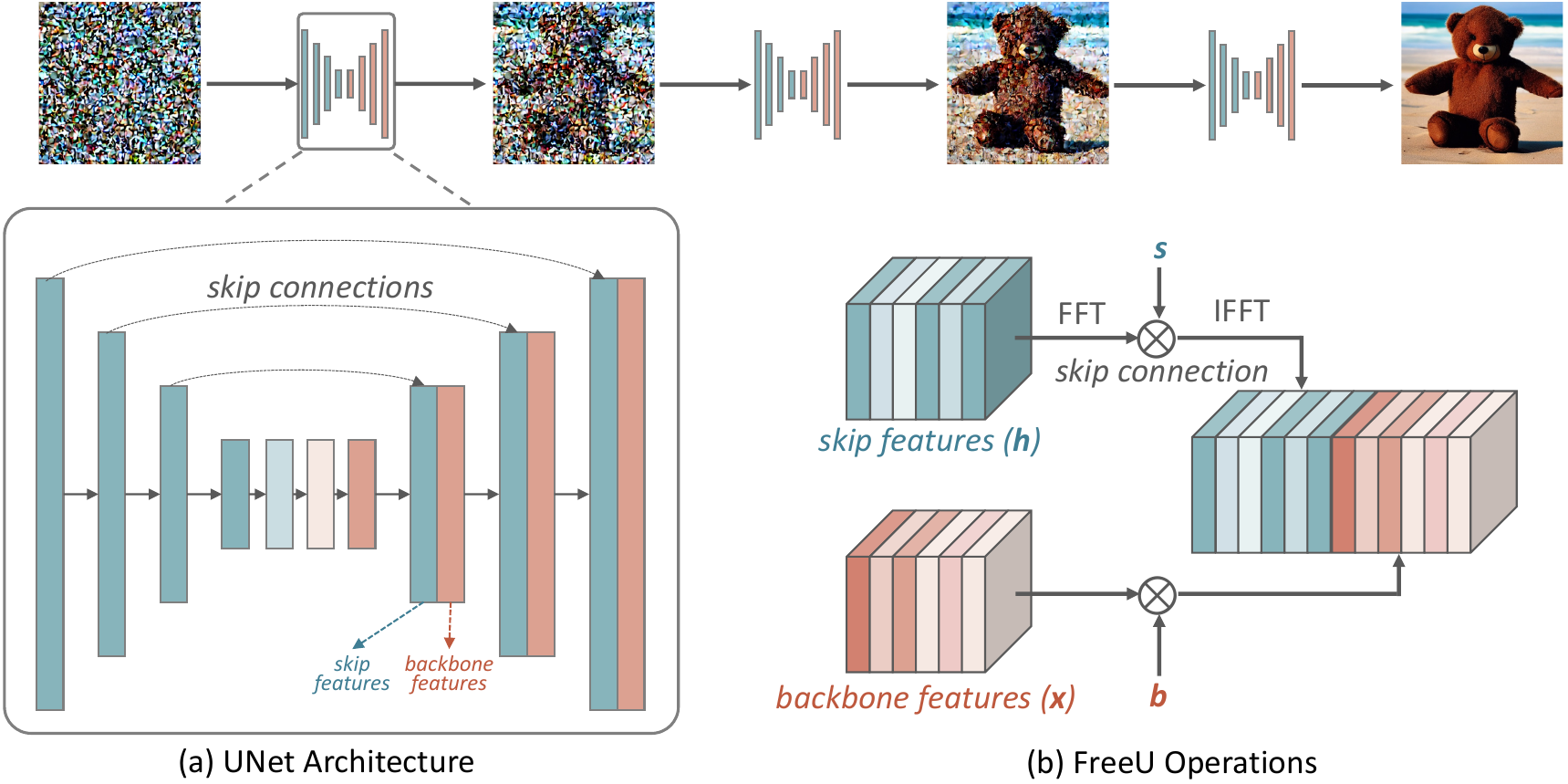}
   \caption{\textbf{FreeU Framework}. \textbf{(a) U-Net Skip Features and Backbone Features}. In U-Net, the skip features and backbone features are concatenated together at each decoding stage. We apply the FreeU operations during concatenation. \textbf{(b) FreeU Operations.} The factor $b$ aims to amplify the backbone feature map $\boldsymbol{x}$, while factor $s$ is designed to attenuate the skip feature map $\boldsymbol{h}$.
   }
   \label{fig:fig_framework}
\end{figure*}

\section{Methodology}

\begin{figure*}
    \centering
    \begin{minipage}{0.6\textwidth}
        \centering
        \includegraphics[width=0.99\linewidth]{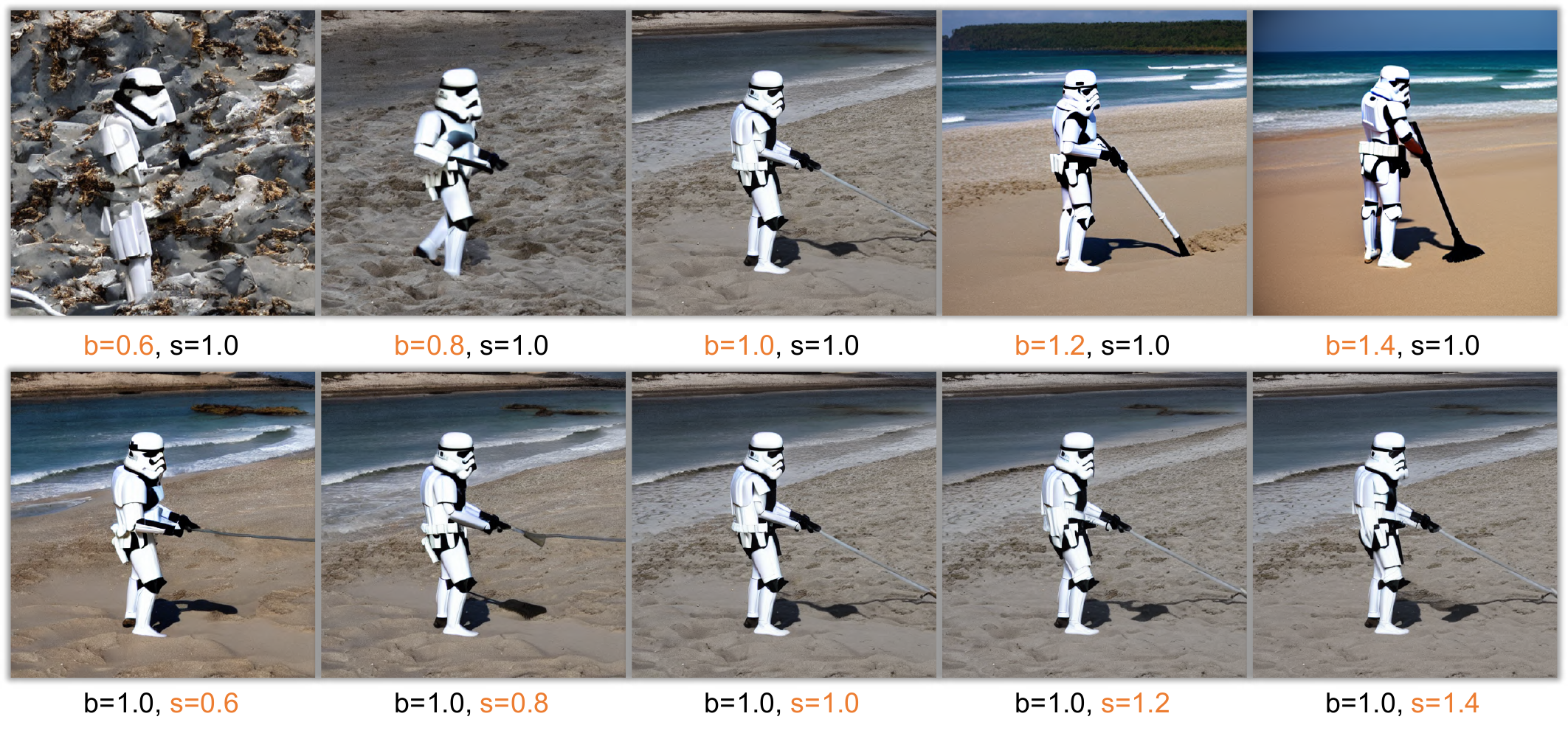}
        \caption{\textbf{Effect of backbone and skip connection scaling factors ($b$ and $s$).}  Increasing the backbone scaling factor $b$ significantly enhances image quality, while variations in the skip scaling factor $s$ have a negligible influence on image synthesis quality.}
        \label{fig:fig_b_s_motivation}
    \end{minipage}\hfill
    \begin{minipage}{0.39\textwidth}
        \centering
        \includegraphics[width=1.0\linewidth]{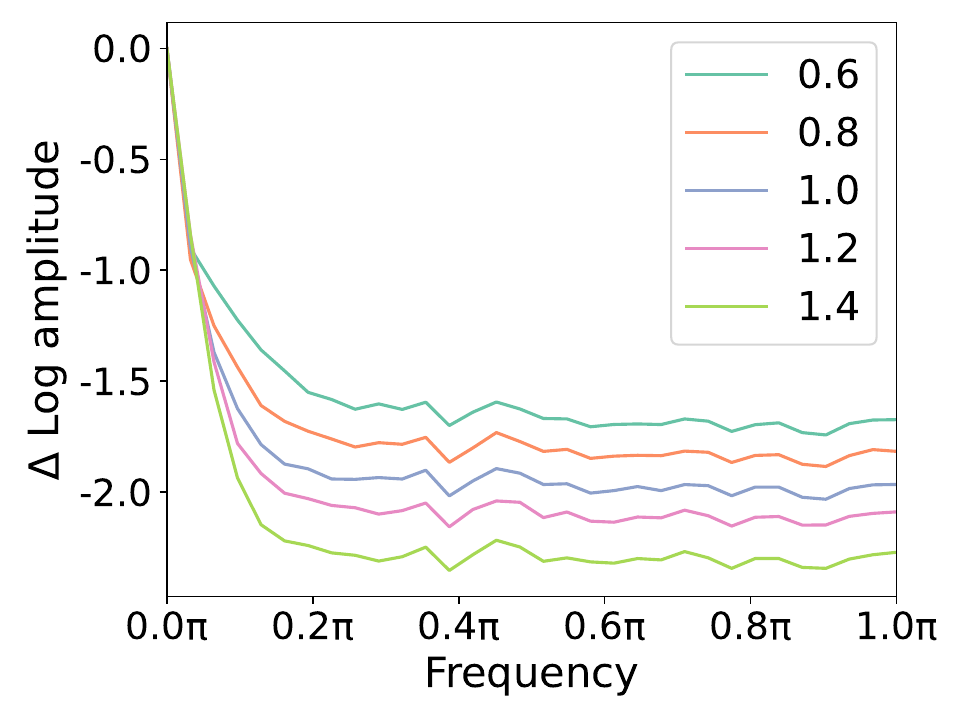}
        \caption{\textbf{Relative log amplitudes of Fourier with variations of the backbone scaling factor $b$.} Increasing in $b$ correspondingly results in a suppression of high-frequency components in the images generated by the diffusion model.}
        \label{fig:fig_bone_factor_compare}
    \end{minipage}
\end{figure*}

\subsection{Preliminaries}
Diffusion models such as Denoising Diffusion Probabilistic Models (DDPM)~\cite{ho2020ddpm}, encompass two fundamental processes for data modeling: a diffusion process and a denoising process. The diffusion process is characterized by a sequence of $T$ steps. At each step $t$, Gaussian noise is incrementally introduced into the data distribution $\boldsymbol{x}_0 \sim q(\boldsymbol{x}_0)$ via a Markov chain, following a prescribed variance schedule denoted as $\beta_1, \ldots, \beta_T$:
\begin{align}
    q(\boldsymbol{x}_t|\boldsymbol{x}_{t-1}) = \mathcal{N}(\boldsymbol{x}_t;\sqrt{1-\beta_t}\boldsymbol{x}_{t-1},\beta_t \mathcal{I}) 
\end{align}
The denoising process reverses the above diffusion process to the underlying clean data $\boldsymbol{x}_{t-1}$ given the noisy input $\boldsymbol{x}_t$:
\begin{align}
    p_\theta(\boldsymbol{x}_{t-1}|\boldsymbol{x}_{t}) = \mathcal{N}(\boldsymbol{x}_{t-1};\boldsymbol{\mu}_\theta(\boldsymbol{x}_{t}, t), \boldsymbol{\Sigma}_\theta(\boldsymbol{x}_{t}, t)) 
\end{align}
The $\boldsymbol{\mu}_\theta$ and $\boldsymbol{\Sigma}_\theta$ determined through estimation procedures involving a denoising model denoted as $\epsilon_\theta$. Typically, this denoising model is implemented using a time-conditional U-Net architecture. It is trained to eliminate noise from data samples while concurrently enhancing the overall fidelity of the generated samples.

\subsection{How does diffusion U-Net perform denoising?}

Building upon the notable disparities observed between low-frequency and high-frequency components throughout the denoising process illustrated in Fig.~\ref{fig:fig_high_low_img_step} and Fig.~\ref{fig:fig_each_step_baseline}, we extend our investigation to delineate the specific contributions of the U-Net architecture within the denoising process, to explore the internal properties of the denoising network. 
As depicted in Fig.~\ref{fig:fig_framework}, the U-Net architecture comprises a primary backbone network, encompassing both an encoder and a decoder, as well as the skip connections that facilitate information transfer between corresponding layers of the encoder and decoder.

\textbf{The backbone of U-Net.}
To evaluate the salient characteristics of the backbone and lateral skip connections in the denoising process, we conduct a controlled experiment wherein we introduce two multiplicative scaling factors—denoted as $b$ and $s$—to modulate the feature maps generated by the backbone and skip connections, respectively, prior to their concatenation. As shown in Fig.~\ref{fig:fig_b_s_motivation},  it is evident that elevating the scale factor $b$ of the backbone distinctly enhances the quality of generated images. Conversely, variations in the scaling factor $s$, which modulates the impact of the lateral skip connections, appear to exert a negligible influence on the quality of the generated images.

Building upon these observations, we subsequently probed the underlying mechanisms that account for the enhancement in image generation quality when the scaling factor $b$ associated with the backbone feature maps is augmented. Our analysis reveals that this quality improvement is fundamentally linked to an amplified denoising capability imparted by the U-Net architecture's backbone. As delineated in Fig.~\ref{fig:fig_bone_factor_compare}, a commensurate increase in $b$ correspondingly results in a suppression of high-frequency components in the images generated by the diffusion model. This implies that enhancing backbone features effectively bolsters the denoising capability of the U-Net architecture, thereby contributing to a superior output in terms of both fidelity and detail preservation.

\begin{figure}[t]
  \centering
   \includegraphics[width=0.9\linewidth]{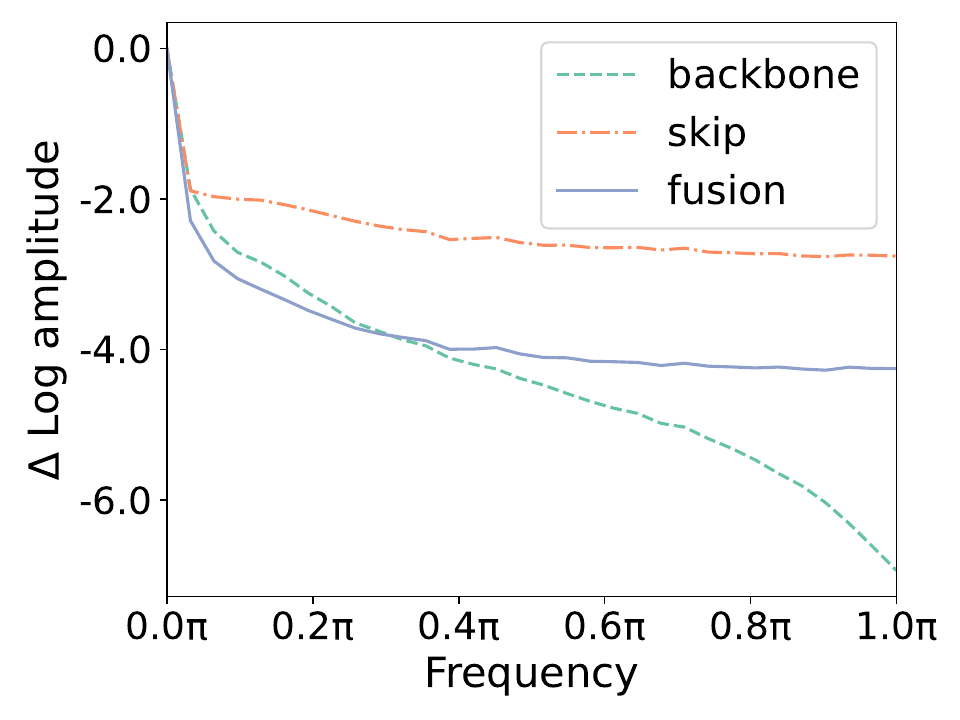}
   \caption{\textbf{Fourie relative log amplitudes of backbone, skip, and their fused feature maps.} The features, forwarded by skip connections directly from earlier layers of the encoder block to the decoder contain a large amount of high-frequency information.}
   \label{fig:fig_skip_bone_fuse}
\end{figure}

\begin{figure}[t]
  \centering
   \includegraphics[width=0.95\linewidth]{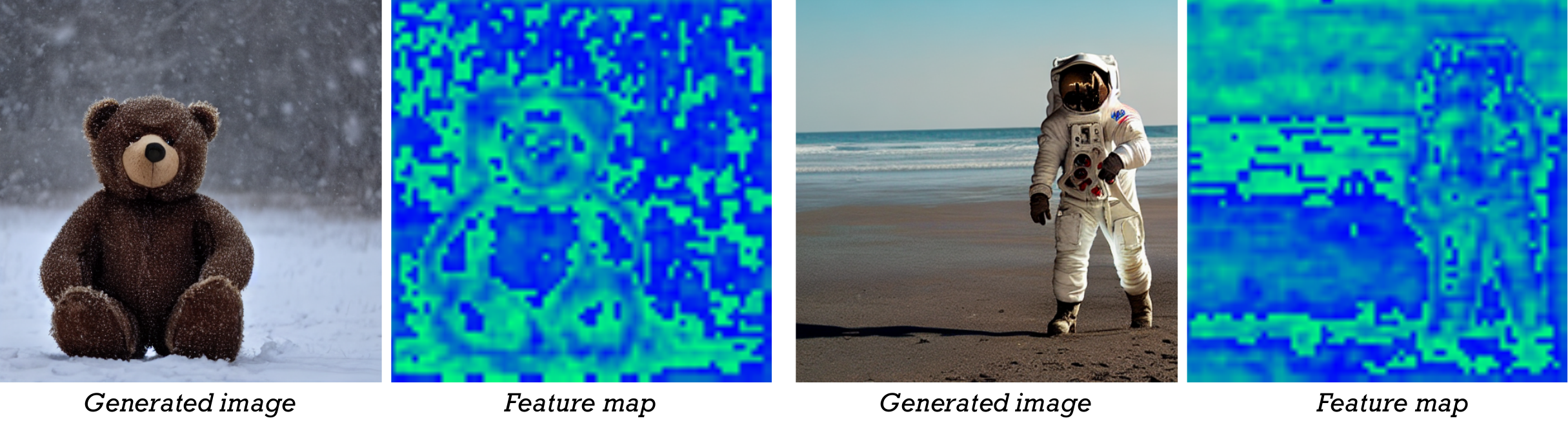}
   \caption{\textbf{Visualization of the average feature maps from the second stage in the decoder}.}
   \label{fig:fig_feature_structure}
\end{figure}

\textbf{The skip connections of U-Net.}
Conversely, the skip connections serve to forward features from the earlier layers of encoder blocks directly to the decoder. Intriguingly, as evidenced in Fig.~\ref{fig:fig_skip_bone_fuse}, these features primarily constitute high-frequency information. Our conjecture, grounded in this observation, posits that during the training of the U-Net architecture, the presence of these high-frequency features may inadvertently expedite the convergence toward noise prediction within the decoder module. Furthermore, the limited impact of modulating skip features in Fig.~\ref{fig:fig_b_s_motivation} also indicates that the skip features predominantly contribute to the decoder's information. This phenomenon, in turn, could result in an unintended attenuation of the efficacy of the backbone's intrinsic denoising capabilities during inference. Thereby, this observation prompts pertinent questions about the counterbalancing roles played by the backbone and the skip connections in the composite denoising performance of the U-Net framework.

\subsection{Free lunch in diffusion U-Net}

Capitalizing on the above discovery, we propel forward with the introduction of simple yet effective method, denoted as ``\textbf{FreeU}'', which effectively bolsters the denoising capability of the U-Net architecture by leveraging the strengths of both components of the U-Net architecture. It substantially improves the generation quality without requiring additional training or fine-tuning.

Technically, for the $l$-th block of the U-Net decoder, let $\boldsymbol{x}_{l}$  represent the backbone feature map from the main backbone at the preceding block, and let $\boldsymbol{h}_{l}$ denote the feature map propagated through the corresponding skip connection. To modulate these feature maps, we introduce two scalar factors: a backbone feature scaling factor $b_l$ for $\boldsymbol{x}_l$ and a yet-to-be-defined skip feature scaling factor $s_l$ for $\boldsymbol{h}_{l}$. Specifically, the factor $b_l$ aims to amplify the backbone feature map $\boldsymbol{x}_l$, while factor $s_l$ is designed to attenuate 
the skip feature map $\boldsymbol{h}_{l}$.

For the backbone features, we introduce a novel method known as structure-related scaling, which dynamically adjusts the scaling of backbone features for each sample. Unlike a fixed scaling factor applied uniformly to all samples or positions within the same channel, our approach adjusts the scaling factor adaptively based on the specific characteristics of the sample features. We first computer the average feature map along the channel dimension:
\begin{gather}
    \boldsymbol{\Bar{x}}_{l} = \frac{1}{C} \sum_{i=1}^C  \boldsymbol{x}_{l,i},
\end{gather}
where $\boldsymbol{x}_{l,i}$ represents the $i$-th channel of the feature map $\boldsymbol{x}_{l}$. $C$ denotes the total number of channels in $\boldsymbol{x}_{l}$. Subsequently, the backbone factor map is determined as follows:
\begin{gather}
    \boldsymbol{\alpha}_{l} = (b_l - 1) \cdot \cfrac{\boldsymbol{\Bar{x}}_{l} - Min(\boldsymbol{\Bar{x}}_{l})}{ Max(\boldsymbol{\Bar{x}}_{l}) - Min(\boldsymbol{\Bar{x}}_{l}) } + 1,
\end{gather}
where $\boldsymbol{\alpha}_{l}$ represents the backbone factor map. $b_l$ is a scalar constant.
Then, upon experimental investigation, we discern that indiscriminately amplifying all channels of $\boldsymbol{x}_l$ through multiplication with $\boldsymbol{\alpha}_{l}$ engenders an oversmoothed texture in the resulting synthesized images. The reason is the enhanced U-Net compromises the image's high-frequency details while denoising. Consequently, we confine the scaling operation to the half channels of $\boldsymbol{x}_l$ as follows:
\begin{align}
    \boldsymbol{x}_{l,i}^{'} = \begin{cases}
    \boldsymbol{x}_{l,i} \odot \boldsymbol{\alpha}_{l}, & \mbox{if } i < C/2\\
    \boldsymbol{x}_{l,i}, &    \mbox{otherwise}
    \end{cases}
\end{align}
Indeed, as illustrated in Fig.~\ref{fig:fig_feature_structure}, the average feature map $\boldsymbol{\Bar{x}}_{l}$ inherently contains valuable structural information. Consequently, the backbone factor map $\boldsymbol{\alpha}_{l}$ is instrumental in amplifying the backbone feature map $\boldsymbol{x}_{l}$ in a manner that aligns with its structural characteristics. This strategic approach serves to mitigate the issue of oversmoothing.
Importantly, this strategy offers a dual benefit. Firstly, it enhances the denoising capabilities of the backbone feature map, allowing it to filter out noise more effectively. Secondly, it avoids the adverse effects associated with the indiscriminate application of scaling across the entire feature map, thereby achieving a more nuanced equilibrium between noise reduction and texture preservation.

\begin{figure*}[ht]
  \centering
   \includegraphics[width=0.99\textwidth]{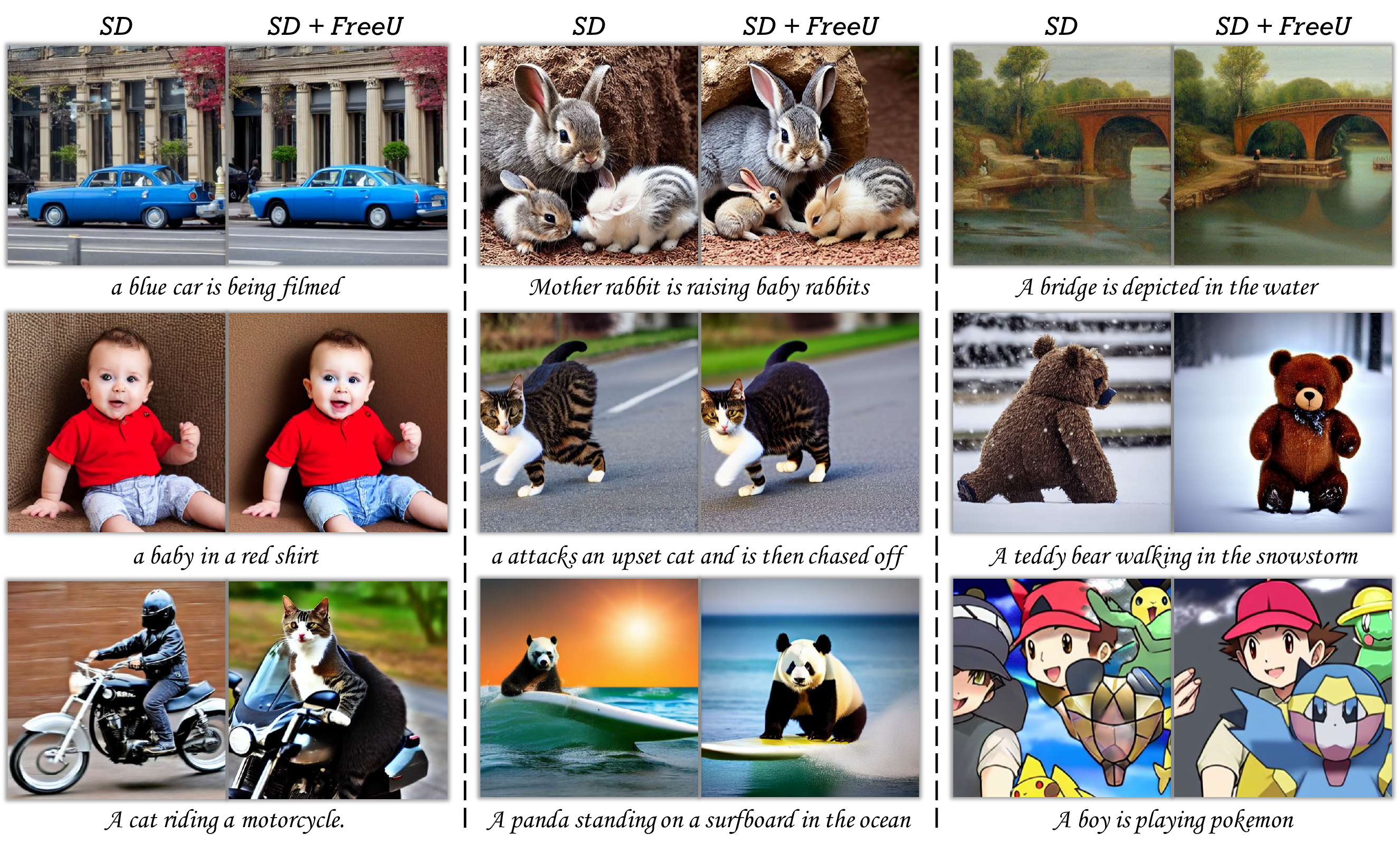}
   \caption{\textbf{Samples generated by Stable Diffusion~\cite{rombach2022ldm} with or without FreeU.}}
   \label{fig:fig_SD_results}
\end{figure*}


To further mitigate the issue of oversmoothed texture due to enhancing denoising, 
we further employ spectral modulation in the Fourier domain to selectively diminish low-frequency components for the skip features. Mathematically, this operation is performed as follows:
\begin{align}
    \boldsymbol{\mathcal{F}}(\boldsymbol{h}_{l,i}) &= \text{FFT}(\boldsymbol{h}_{l,i}) \\
    \boldsymbol{\mathcal{F}}'(\boldsymbol{h}_{l,i}) &= \boldsymbol{\mathcal{F}}(\boldsymbol{h}_{l,i}) \odot \boldsymbol{\beta}_{l,i} \\
    \boldsymbol{h}_{l,i}' &= \text{IFFT}(\boldsymbol{\mathcal{F}}'(\boldsymbol{h}_{l,i}))
\end{align}
where $\text{FFT}(\cdot)$ and $\text{IFFT}(\cdot)$ are Fourier transform and inverse Fourier transform. $\odot$ denotes element-wise multiplication, and $\boldsymbol{\beta}_{l,i}$ is a Fourier mask, designed as a function of the magnitude of the Fourier coefficients, serving to implement the frequency-dependent scaling factor $s_l$:
\begin{align}
\boldsymbol{\beta}_{l,i}(r) = 
    \begin{cases} 
    s_l & \text{if } r < r_{\text{thresh}}, \\
    1 & \text{otherwise}.
    \end{cases}
\end{align}
where $r$ is the radius. $r_{\text{thresh}}$ is the threshold frequency.
Then, the augmented skip feature map $\boldsymbol{h}_{l}'$ is then concatenated with the modified backbone feature map $\boldsymbol{x}_{l}'$ for subsequent layers in the U-Net architecture, as shown in Fig.~\ref{fig:fig_framework}.

Remarkably, the proposed FreeU framework does not require any task-specific training or fine-tuning. Adding the backbone and skip scaling factors can be easily done with just a few lines of code. Essentially, the parameters of the architecture can be adaptively re-weighted during the inference phase, which allows for a more flexible and potent denoising operation without adding any computational burden. This makes FreeU a highly practical solution that can be seamlessly integrated into existing diffusion models to improve their performance.

%% file: sec/3_experiment.tex
\section{Experiments}

\begin{figure*}[ht]
  \centering
   \includegraphics[width=0.99\textwidth]{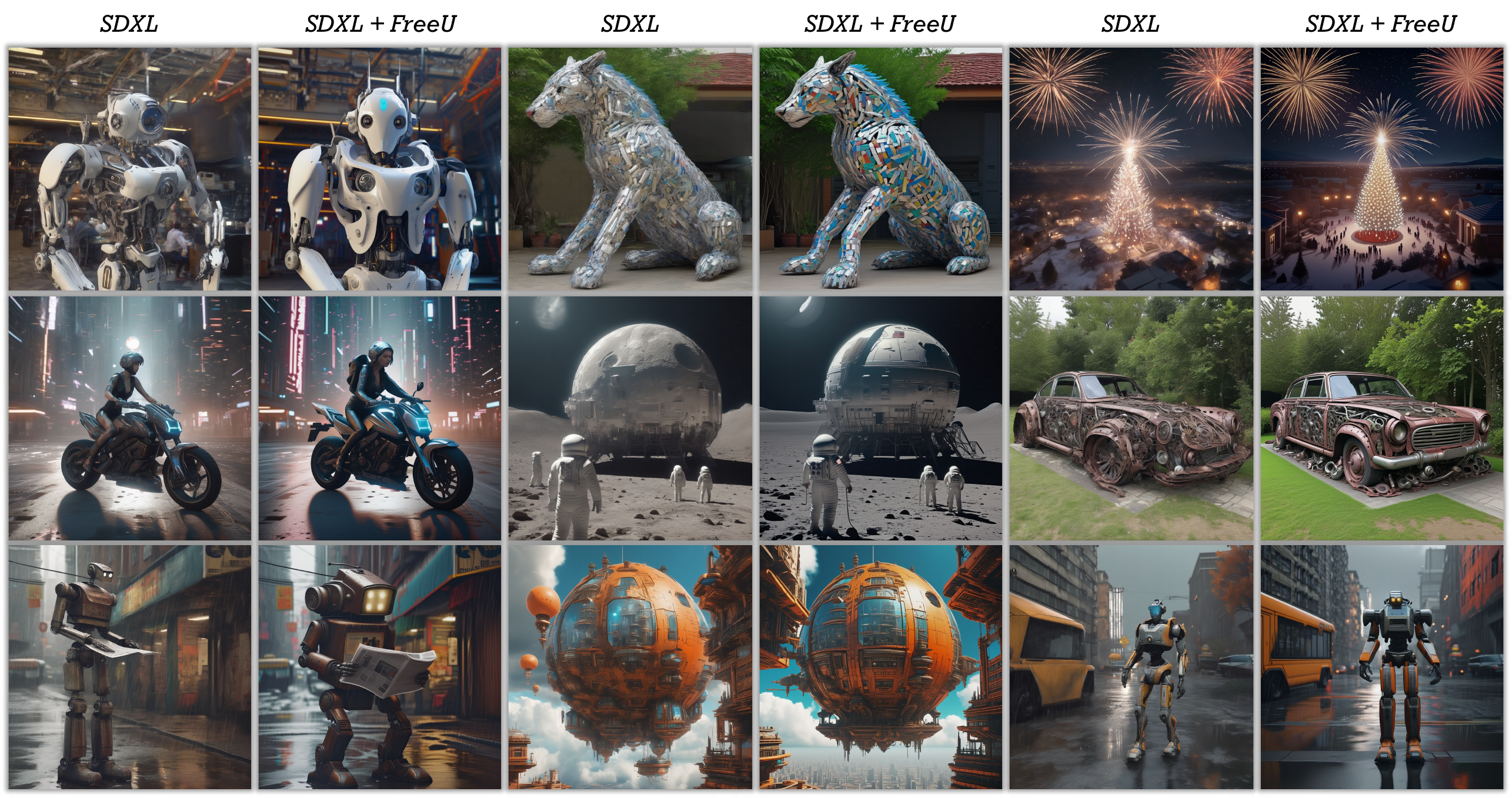}
   \caption{\textbf{Samples generated by Stable Diffusion-XL~\cite{podell2023sdxl} with or without FreeU.}}
   \label{fig:fig_SDXL_results}
\end{figure*}

\subsection{Implementation details}

To assess the effectiveness of the proposed FreeU, we systematically conduct a series of experiments, aligning our benchmarks with state-of-the-art methods such as Stable Diffusion~\cite{rombach2022ldm}, DreamBooth~\cite{ruiz2022dreambooth}, ModelScope~\cite{luo2023videofusion}, and Rerender~\cite{yang2023rerender}. Importantly, our approach seamlessly integrates with these established methods without imposing any additional computational overhead associated with supplementary training or fine-tuning. We meticulously adhere to the prescribed settings of these methods and exclusively introduce the backbone feature factors and skip feature factors during the inference.

\subsection{Text-to-image}

Stable Diffusion~\cite{rombach2022ldm} is a latent text-to-image diffusion model renowned for its capability to generate photorealistic images based on textual input. It has consistently demonstrated exceptional performance in various image synthesis tasks. With the integration of our FreeU augmentation into Stable Diffusion, the results, as exemplified in Fig.~\ref{fig:fig_SD_results}, exhibit a notable enhancement in the model's generative capacity.

\begin{figure*}
  \centering
   \includegraphics[width=0.9\textwidth]{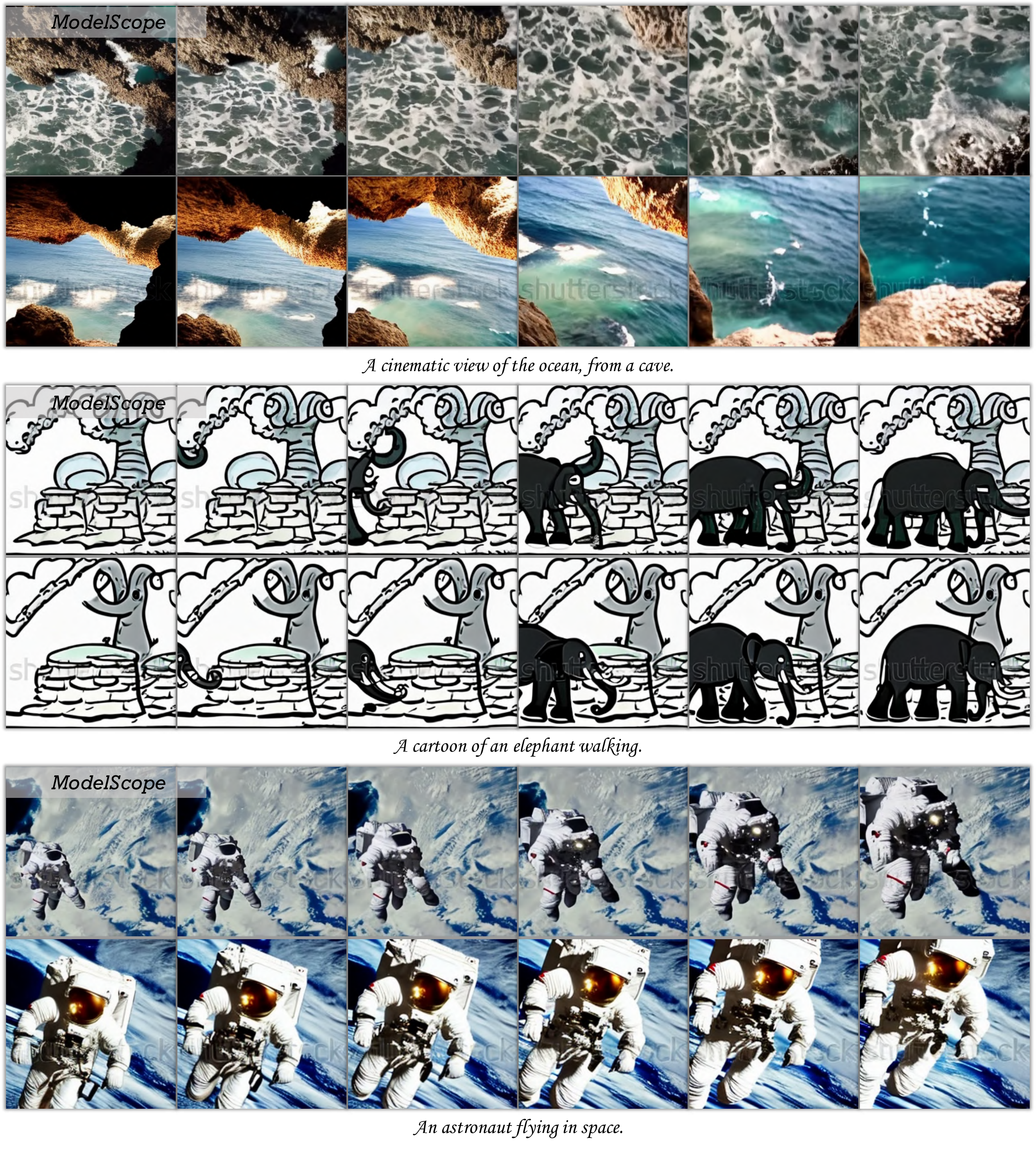}
   \caption{\textbf{Samples generated by ModelScope~\cite{luo2023videofusion} with or without FreeU.}
   }
   \label{fig:fig_fusionvideo}
\end{figure*}

To elaborate, the incorporation of FreeU into Stable Diffusion~\cite{rombach2022ldm} yields improvements in both entity portrayal and fine-grained details. For instance, when provided with the prompt \textit{``a blue car is being filmed''}, FreeU refines the image, eliminating rooftop irregularities and enhancing the textural intricacies of the surrounding structures. In the case of \textit{``Mother rabbit is raising baby rabbits''}, FreeU ensures that the generated image portrays a mother rabbit in a normal appearance caring for baby rabbits. Furthermore, In scenarios like \textit{``a attacks an upset cat and is then chased off''} and \textit{``A teddy bear walking in the snowstorm''}, FreeU helps generate more realistically posed cats and teddy bears. Impressively, in response to the complex prompt \textit{``A cat riding a motorcycle''}, FreeU not only accurately renders the individual entities but also expertly captures the nuanced relationship between them, ensuring that the cat is actively engaged in riding. In Figure~\ref{fig:fig_SDXL_results}, we present the generated images based on the SDXL framework~\cite{podell2023sdxl}. It becomes evident that our proposed FreeU consistently excels in generating realistic images, especially in detail generation. These compelling results serve as a testament to the substantial qualitative enhancements engendered by the synergy of FreeU with the SD\cite{rombach2022ldm} or SDXL\cite{podell2023sdxl} frameworks.

\noindent \textbf{Quantitative evaluation.} 
We conduct a study with 35 participants to assess \textit{image quality} and \textit{image-text alignment}. Each participant receives a text prompt and two corresponding synthesized images, one from SD and another from SD+FreeU. To ensure fairness, we use the same randomly sampled random seed for generating both images. The image sequence is randomized to eliminate any bias.
Participants then select the image they consider superior for \textit{image-text alignment} and \textit{image quality}, respectively. We tabulate the votes for SD and SD+FreeU in each category in Table~\ref{tab:t2i}. Our analysis reveals that the majority of votes go to SD+FreeU, indicating that FreeU significantly enhances the Stable Diffusion text-to-image model in both evaluated aspects.

\begin{table}[t]
  \centering
  \caption{\textbf{Text-to-Image Quantitative Results.} We count the percentage of votes for the baseline and our method respectively. \textit{Image-Text} refers to \textit{Image-Text Alignment}. }
  \vspace{-0.5em}
    \small 
    \begin{tabular}{l|c|c}
    \Xhline{1pt}
    \textbf{Method} & \textbf{Image-Text} & \textbf{Image Quality} \\ \Xhline{1pt}
    SD~\cite{rombach2022ldm} & 14.12\% & 14.66\%  \\ 
    \textbf{SD+FreeU} & \textbf{85.88\%} & \textbf{85.34\%} \\
    \Xhline{1pt}
  \end{tabular}
  \label{tab:t2i}
\end{table}

\begin{table}[t]
  \centering
  \caption{\textbf{Text-to-Video Quantitative Results.} We count the percentage of votes for the baseline and our method respectively. \textit{Video-Text} refers to \textit{Video-Text Alignment}. }
  \vspace{-0.5em}
    \small 
    \begin{tabular}{l|c|c}
    \Xhline{1pt}
    \textbf{Method} & \textbf{Video-Text} & \textbf{Video Quality} \\ \Xhline{1pt}
    ModelScope~\cite{luo2023videofusion} & 15.29\% & 14.33\%  \\ 
    \textbf{ModelScope+FreeU} & \textbf{84.71\%} & \textbf{85.67\%} \\
    \Xhline{1pt}
  \end{tabular}
  \label{tab:t2v}
\end{table}

\begin{figure}[t]
  \centering
\includegraphics[width=0.49\textwidth]{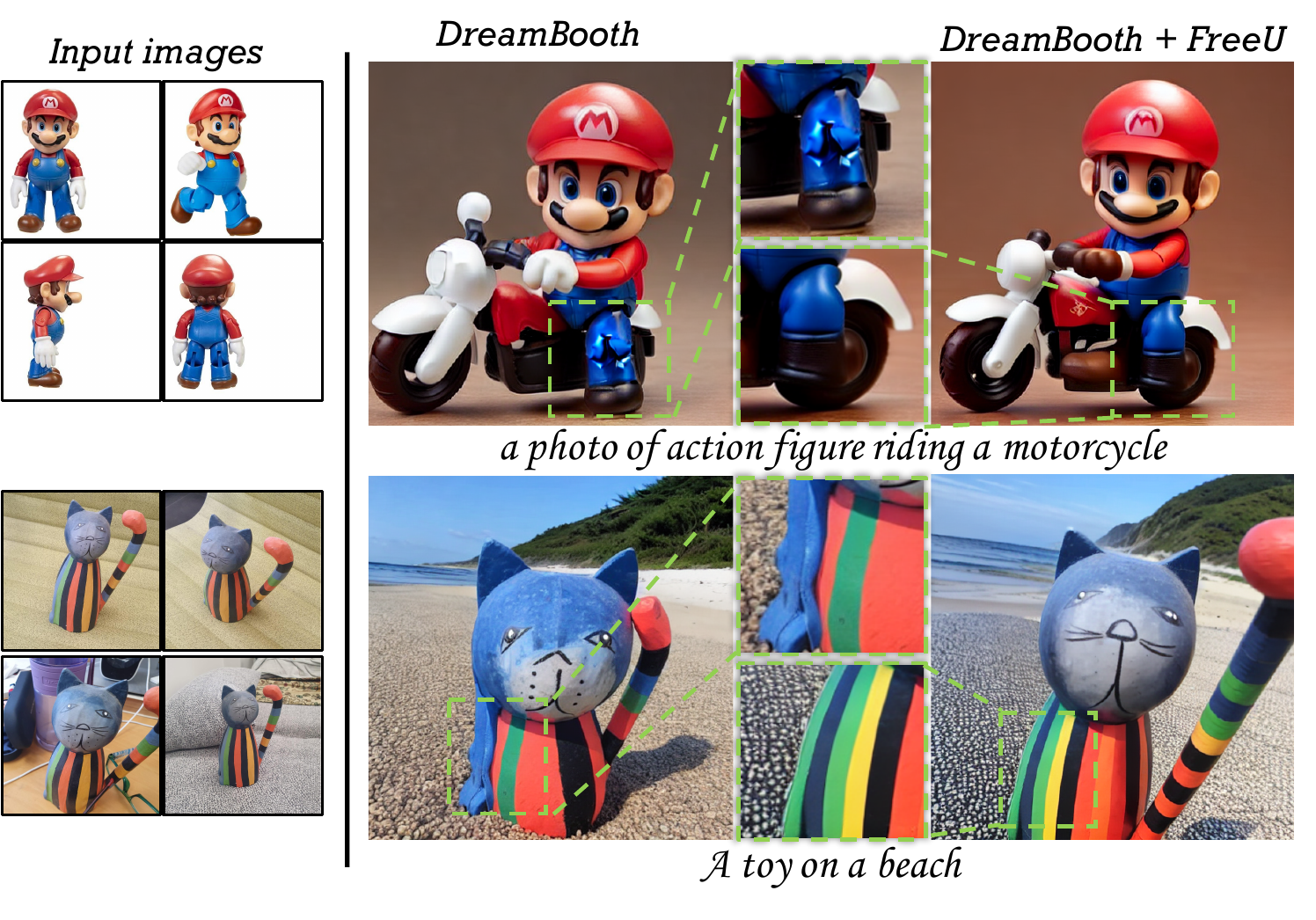}
   \caption{\textbf{Samples generated by DreamBooth~\cite{ruiz2022dreambooth} with or without FreeU.}}
   \label{fig:fig_dreambooth}
\end{figure}

\subsection{Text-to-video}

ModelScope~\cite{luo2023videofusion}, an avant-garde text-to-video diffusion model, stands at the forefront of video generation from textual descriptions. The infusion of our FreeU augmentation into ModelScope~\cite{luo2023videofusion} serves to further hone its video synthesis prowess, as substantiated by Fig.~\ref{fig:fig_fusionvideo}. For instance, when presented with the prompt \textit{``A cinematic view of the ocean, from a cave''}, FreeU enables ModelScope~\cite{luo2023videofusion} to generate the perspective ``from a cave'', enriching the visual narrative. In the case of \textit{``A cartoon of an elephant walking''}, ModelScope~\cite{luo2023videofusion} initially generates an elephant with two trunks, but with the incorporation of FreeU, it rectifies this anomaly and produces a correct depiction of an elephant in motion. Moreover, in response to the prompt \textit{``An astronaut flying in space''}, ModelScope~\cite{luo2023videofusion}, with the assistance of FreeU, can generate a clear and vivid portrayal of an astronaut floating in the expanse of outer space.

\begin{figure}[t]
  \centering
   \includegraphics[width=0.49\textwidth]{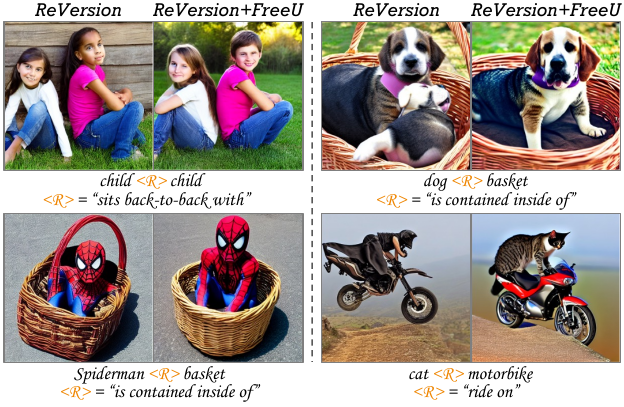}
   \caption{\textbf{Samples generated by ReVersion~\cite{huang2023reversion} with or without FreeU.}}
   \label{fig:fig_reversion}
\end{figure}

\begin{figure}[t]
  \centering
   \includegraphics[width=0.49\textwidth]{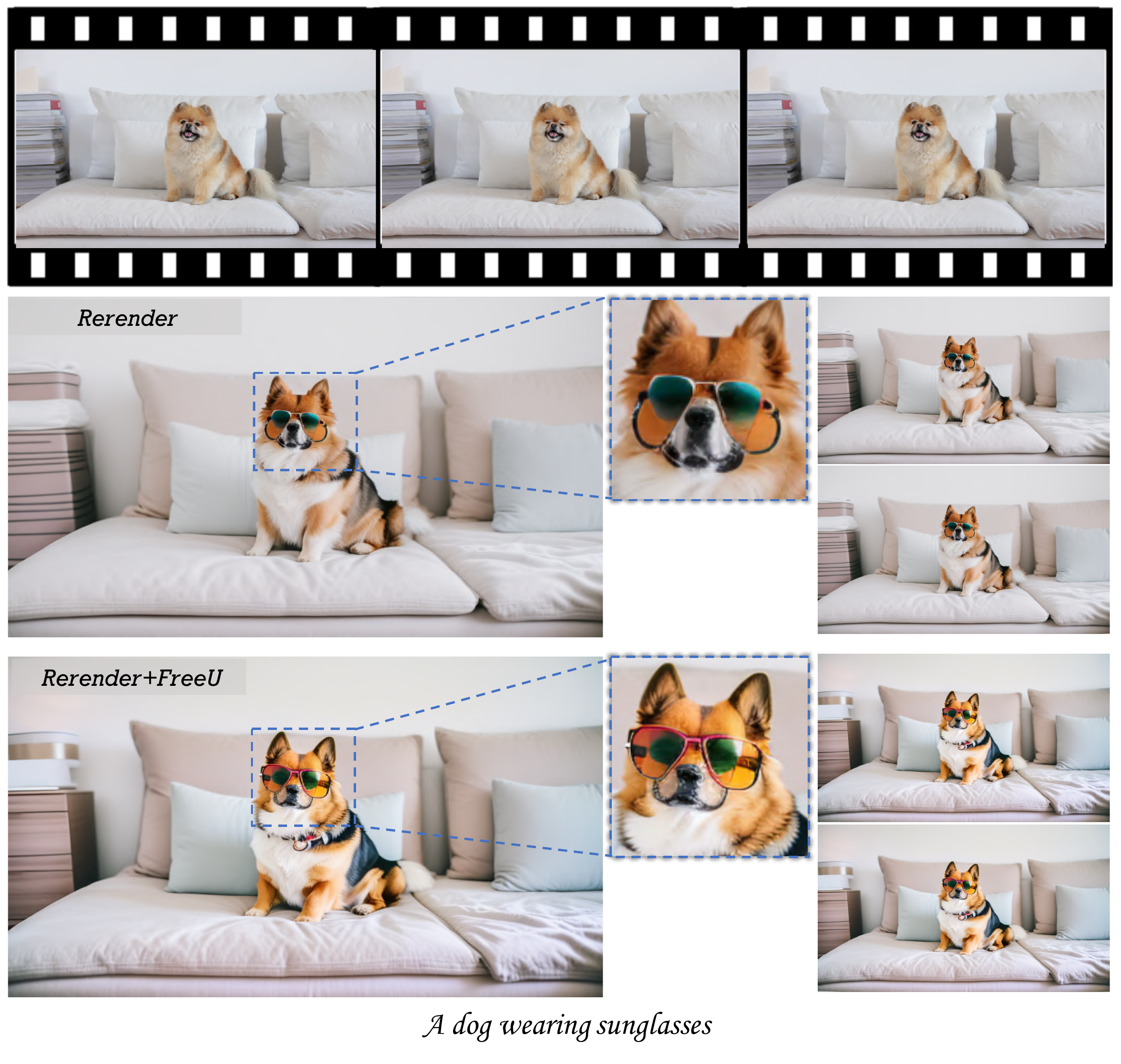}
   \caption{\textbf{Samples generated by Rerender~\cite{yang2023rerender} with or without FreeU.}}
   \label{fig:fig_Rerender}
\end{figure}

These results underscore the significant improvements achieved through the synergistic application of FreeU with ModelScope~\cite{luo2023videofusion}, resulting in high-quality generated content characterized by clear motion, rich detail, and semantic alignment.

\begin{figure*}[t]
  \centering
   \includegraphics[width=0.99\textwidth]{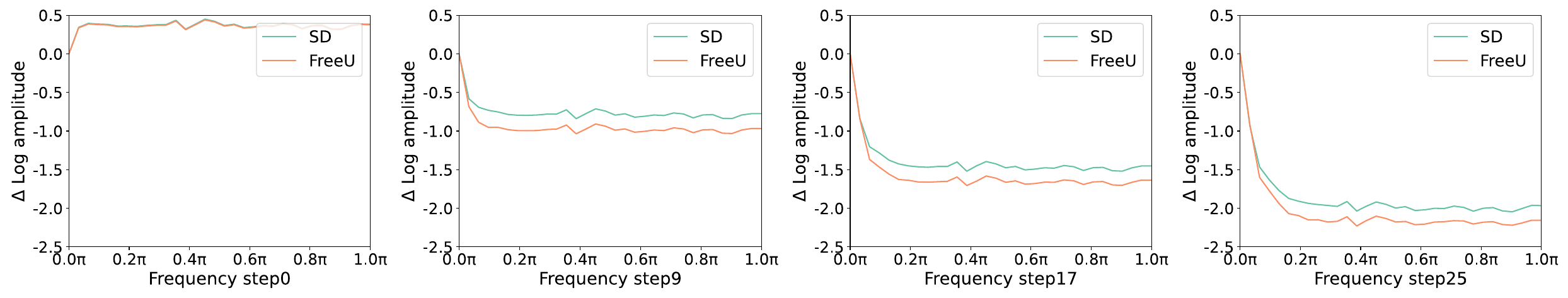}
   \caption{\textbf{Fourier relative log amplitudes of Stable Diffusion~\cite{rombach2022ldm} with or without FreeU within the denoising process.}
   }
   \label{fig:fft_each_step_compare}
\end{figure*}

\begin{figure*}[t]
  \centering
   \includegraphics[width=0.99\textwidth]{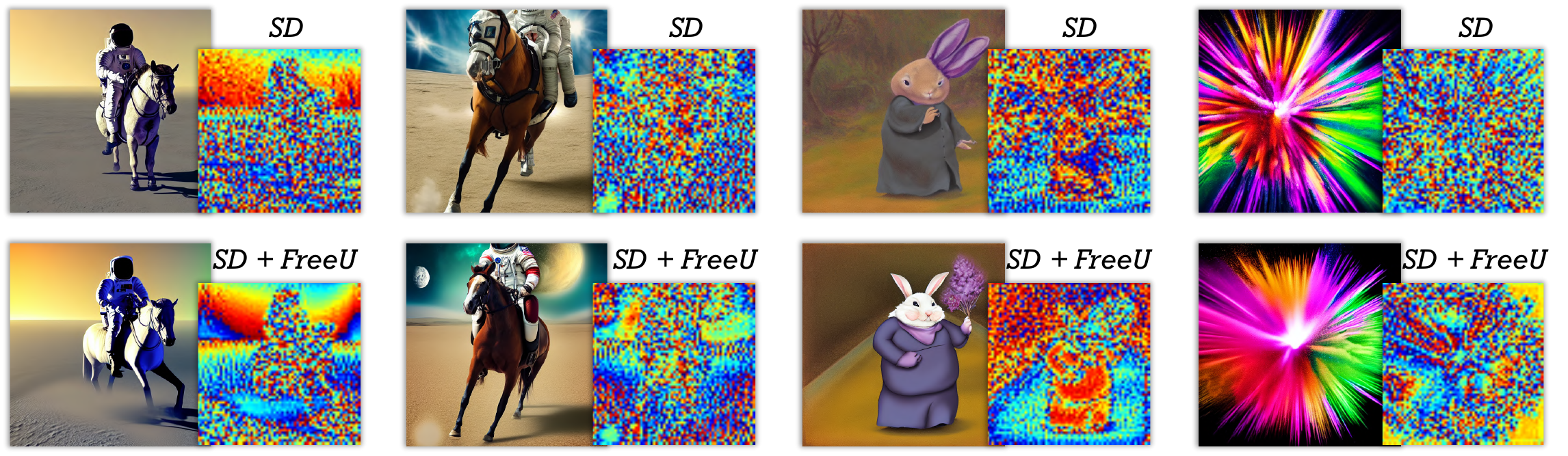}
   \caption{The visualization of feature maps for Stable Diffusion~\cite{rombach2022ldm} with or without FreeU.
   }
   \label{fig:fig_feature_visualize}
\end{figure*}

\noindent \textbf{Quantitative evaluation.} 
We conduct the quantitative evaluation for FreeU on the text-to-video task in a similar way as text-to-image. The results displayed in Table~\ref{tab:t2v} indicate that most participants prefer the video generated with FreeU.

\subsection{Downstream tasks}

FreeU presents substantial enhancements in the quality of synthesized samples across various diffusion model applications. Our evaluations extend from foundational image and video synthesis models to more specialized downstream applications.

We incorporate FreeU into Dreambooth~\cite{ruiz2022dreambooth}, a diffusion model specialized in personalized text-to-image tasks. The enhancements are evident, as demonstrated in Fig.~\ref{fig:fig_dreambooth}, the synthesized images present marked improvements in realism. For instance, while the base DreamBooth~\cite{ruiz2022dreambooth} model struggles to synthesize the appearance of the action figure's legs from the prompt \textit{``a photo of action figure riding a motorcycle''}, the FreeU-augmented version deftly overcomes this hurdle. Similarly, for the prompt \textit{``A toy on a beach''}, the initial output exhibited body shape anomalies. FreeU's integration refines these imperfections, providing a more accurate representation and improving color fidelity.

We also integrate FreeU into ReVersion~\cite{huang2023reversion}, a Stable Diffusion based relation inversion method, enhancing its quality as shown in Fig.~\ref{fig:fig_reversion}. For example, when the relation ``back to back'' is to be expressed between two children, FreeU enhances ReVersion's ability to accurately represent this relationship. For the ``inside'' relation, when a \textit{dog} is supposed to be placed inside of a \textit{basket}, ReVersion sometimes generates a dog with artifacts, and introducing FreeU helps eliminate these artifacts. While ReVersion effectively captures relational concepts, Stable Diffusion might occasionally struggle to synthesize the relation concept due to excessive high-frequency noises in the U-Net skip features. Adding FreeU allows better entity and relation synthesis quality by using exactly the same relation prompt learned by ReVersion. 

Furthermore, we evaluated FreeU's impact on Rerender~\cite{yang2023rerender}, a diffusion model tailored for zero-shot text-guided video-to-video translations. Fig.~\ref{fig:fig_Rerender} depicts the results: clear improvements in the detail and realism of synthesized videos. For instance, when provided with the prompt \textit{``A dog wearing sunglasses''} and an input video, Rerender~\cite{yang2023rerender} initially produces a dog video with artifacts related to the \textit{``sunglasses''}. However, the incorporation of FreeU successfully eliminates such artifacts, resulting in a refined output.

In summation, these outcomes substantiate that the incorporation of FreeU leads to enhanced entity representation and synthesis quality, employing precisely the same learned prompt.

\begin{figure}[t]
  \centering
   \includegraphics[width=0.49\textwidth]{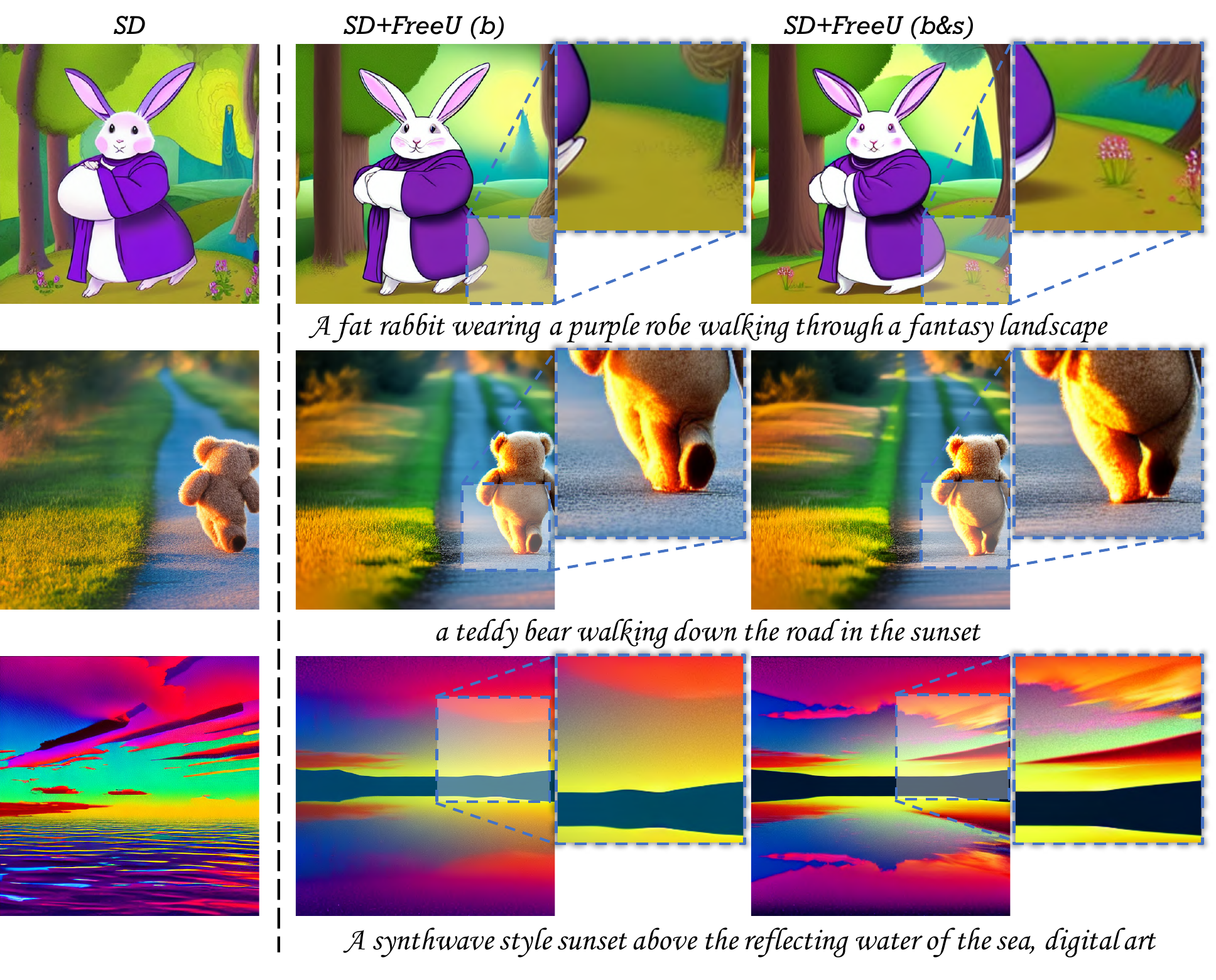}
   \caption{\textbf{The ablation study of backbone scaling factor and skip scaling factor.}}
   \label{fig:fig_ablation_framework}
\end{figure}

\begin{figure}[ht]
  \centering
   \includegraphics[width=0.5\textwidth]{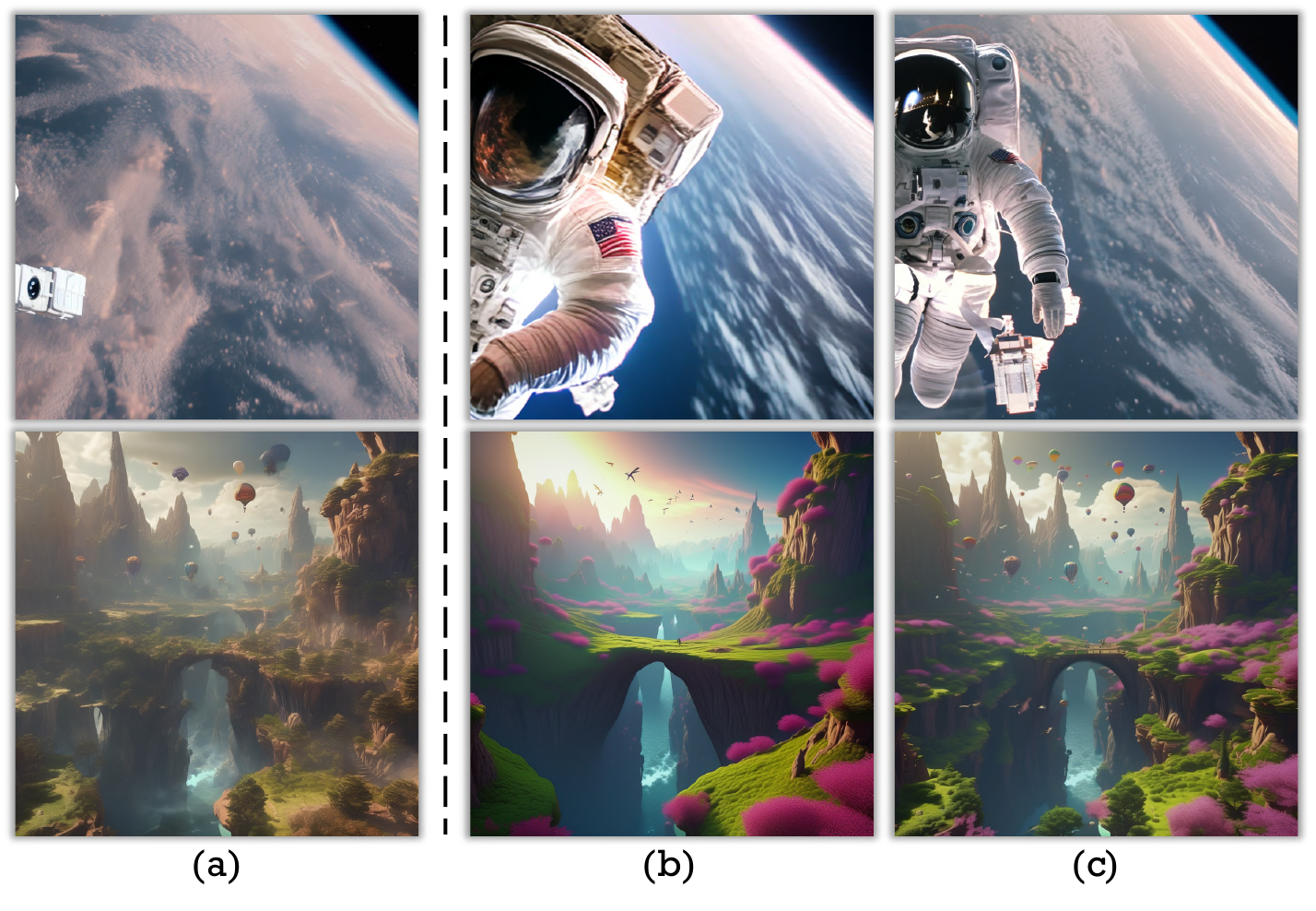}
   \caption{\textbf{The ablation study of backbone scaling factor. (a) The generated images of SD. (b) The generated images of FreeU with a constant factor. (c) The generated images of FreeU with the structure-related scaling factor map. 
}}
   \label{fig:fig_backbone_factor_v1_v2_ablation}
\end{figure}

\subsection{Ablation study}

\textbf{Effects of FreeU.}
FreeU is introduced with the primary aim of enhancing the denoising capabilities of the U-Net architecture within the diffusion model. To assess the impact of FreeU, we conducted analytical experiments using Stable Diffusion~\cite{rombach2022ldm} as the base framework. In Fig.~\ref{fig:fft_each_step_compare}, we present visualizations of the relative log amplitudes of the Fourier transform of Stable Diffusion~\cite{rombach2022ldm}, comparing cases with and without the incorporation of FreeU. These visualizations illustrate that FreeU exerts a discernible influence in reducing high-frequency information at each step of the denoising process, which indicates FreeU's capacity to effectively denoising. Furthermore, we extended our analysis by visualizing the feature maps of the U-Net architecture. As shown in Fig.~\ref{fig:fig_feature_visualize}, we observe that the feature maps generated by FreeU contain more pronounced structural information. This observation aligns with the intended effect of FreeU, as it preserves intricate details while effectively removing noise, harmonizing with the denoising objectives of the model.

\textbf{Effects of components in FreeU.} We evaluate the effects of the proposed FreeU strategy, \ie introducing backbone feature scaling factors and skip feature scaling factors to intricately balance the feature contributions from the U-Net architecture’s primary backbone and skip connections. In Fig.~\ref{fig:fig_ablation_framework}, we present the results of our evaluations. In the case of \textit{SD+FreeU(b)}, where backbone scaling factors are integrated during inference, we observe a noticeable improvement in the generation of vivid details compared to \textit{SD}~\cite{rombach2022ldm} alone. For instance, when given the prompt \textit{``A fat rabbit wearing a purple robe walking through a fantasy landscape''}, \textit{SD+FreeU(b)} generates a more realistic rabbit with normal arms and ears, as opposed to \textit{SD}~\cite{rombach2022ldm}. However, it is imperative to note that while the inclusion of feature scaling factors yields significant improvements, it can occasionally lead to an undesirable oversmoothing of textures. To mitigate this issue, we introduce skip feature scaling factors, aiming to reduce low-frequency information and alleviate the problem of texture oversmoothing. As demonstrated in Fig.~\ref{fig:fig_ablation_framework}, the combination of both backbone and skip feature scaling factors in \textit{SD+FreeU(b \& s)} leads to the generation of more realistic images. For instance, in the prompt \textit{``A synthwave style sunset above the reflecting water of the sea, digital art''}, the generated sunset sky in \textit{SD+FreeU(b \& s)} exhibits enhanced realism compared to \textit{SD+FreeU(b)}. This highlights the efficacy of the comprehensive FreeU strategy in balancing features and mitigating issues related to texture smoothing, ultimately resulting in more faithful and realistic image generation.

\textbf{Effects of backbone structure-related factor.}
We evaluate the effects of the proposed backbone scaling strategy, structure-related scaling, on the delicate balance between noise reduction and texture preservation. Illustrated in Figure~\ref{fig:fig_backbone_factor_v1_v2_ablation}, when compared to the results generated by \textit{SD}~\cite{rombach2022ldm}, we observe a substantial enhancement in the image quality generated by FreeU when utilizing a constant scaling factor. However, it is pertinent to highlight that the utilization of a fixed scaling factor can engender adverse consequences, manifesting as pronounced oversmoothing of textures and undesirable color oversaturation. Conversely, FreeU with the structure-related scaling factor map employs an adaptive scaling approach, leveraging structural information to guide the assignment of the backbone factor map. Our observations indicate that FreeU with the structure-related scaling factor map effectively mitigates these issues and achieves significant improvements in generating vivid and intricate details.

%% file: sec/4_conclusion.tex
\section{Conclusion}

In this study, we introduce the elegantly simple yet highly effective approach, termed \textbf{\textit{FreeU}}, which substantially enhances the sample quality of diffusion models without incurring any additional computational costs. Motivated by the fundamental role played by both skip connections and backbone features in U-Net architectures, we conduct an in-depth analysis of their effects in diffusion U-Net. Our investigation reveals that the primary backbone primarily contributes to denoising, while the skip connections predominantly introduce high-frequency features into the decoder, potentially leading to a neglect of essential backbone semantics. To address this, we strategically re-weight the contributions originating from the U-Net’s skip connections and backbone feature maps. This re-weighting process capitalizes on the unique strengths of both U-Net components, resulting in a substantial improvement in sample quality across a wide range of text prompts and random seeds. Our proposed \textbf{\textit{FreeU}} can be seamlessly integrated into various diffusion foundation models and their downstream tasks, offering a versatile means of enhancing sample quality.